\documentclass[sigconf]{acmart}

\AtBeginDocument{%
  }

\setcopyright{none}
\renewcommand\footnotetextcopyrightpermission[1]{}
\copyrightyear{}
\acmYear{2026}
\acmDOI{}
\acmISBN{}
\acmPrice{}

\acmConference[ICMHI 2026]{10th International Conference on Medical and Health Informatics (ICMHI 2026)}{May 15--17, 2026}{Kyoto, Japan}

\begin{document}

\title{Benchmarking Convolutional, Transformer, Hybrid, and Vision Language Models for Multi Disease Retinal Screening}

\author{Durjoy Dey}
\email{durjoy.dey@mail.concordia.ca}
\affiliation{%
  \department{Department of Computer Science and Software Engineering}
  \institution{Concordia University}
  \streetaddress{1455 Blvd. De Maisonneuve Ouest}
  \city{Montreal}
  \state{QC}
  \postcode{H3G 1M8}
  \country{Canada}
}
\affiliation{%
  \institution{Ebovir Biotechnologie Inc.}
  \city{Montreal}
  \state{QC}
  \country{Canada}
}

\author{Aymane Ajbar}
\email{a_ajbar@live.concordia.ca}
\affiliation{%
  \department{Department of Computer Science and Software Engineering}
  \institution{Concordia University}
  \streetaddress{1455 Blvd. De Maisonneuve Ouest}
  \city{Montreal}
  \state{QC}
  \country{Canada}
}
\affiliation{%
  \institution{Ebovir Biotechnologie Inc.}
  \city{Montreal}
  \state{QC}
  \country{Canada}
}

\author{Yuhong Yan}
\email{yuhong.yan@concordia.ca}
\affiliation{%
  \department{Department of Computer Science and Software Engineering}
  \institution{Concordia University}
  \streetaddress{1455 Blvd. De Maisonneuve Ouest}
  \city{Montreal}
  \state{QC}
  \country{Canada}
}

\renewcommand{\shortauthors}{Durjoy Dey et al.}

\begin{abstract}
Modern deep learning offers powerful tools for automated retinal screening, but it remains unclear how different visual model families compare in realistic multi-disease settings and under domain shift. In this work, we benchmark twelve architectures across four model families: convolutional neural networks (CNNs), vision transformers (ViTs), hybrid CNN–transformer backbones, and vision–language models (VLMs), using the Retinal Fundus Multi-disease Image Dataset (RFMiD), a multi-label fundus dataset with 28 disease classes. We evaluate two tasks relevant to screening practice: (i) a binary screening task that determines whether an image contains any retinal disease, and (ii) a multi-label classification task, which predicts the full spectrum of RFMiD pathologies present in each image. Using standardized family-specific training configurations and a unified calibration and evaluation protocol, we report AUC, F1, precision, recall, and sensitivity at a clinically relevant operating point with specificity near 80\% across all models. On RFMiD, all architectures perform well on the binary screening task (AUC above 84\%), but attention-based models dominate: SwinTiny and the hybrid CoAtNet0 and MaxViTTiny achieve the highest AUC and F1, and significantly improve macro and micro F1 in the more challenging multi-label setting. Vision–language models (CLIP ViT-B/16 and SigLIP-Base384) are competitive with CNN baselines but do not surpass the best transformer and hybrid backbones. In external validation on Messidor-2 for referable diabetic retinopathy (DR), AUC ranges from 66.8\% to 84.7\%, with hybrid and transformer models again showing the strongest performance. The SigLIP-Base384 vision–language model demonstrates favorable precision–recall trade-offs for referable DR. To our knowledge, this is the first study to benchmark convolutional, transformer, hybrid, and vision–language models on RFMiD using a unified evaluation protocol with external validation on Messidor-2. Our results provide a reproducible reference for model choice in multi-disease retinal screening and guide the development of future automated screening tools for clinical deployment.
\end{abstract}





\thanks{Accepted at ICMHI 2026, 10th International Conference on Medical and Health Informatics, Kyoto, Japan. To appear in ACM Conference Proceedings.}

\maketitle

\section{Introduction}

The World Health Organization reports that at least 2.2 billion people globally live with near or distance vision impairment, and in at least 1 billion of these cases the impairment could have been prevented or remains unaddressed \cite{WHO2023Blindness}. The leading causes of vision impairment and blindness include uncorrected refractive errors, cataract, and retinal diseases such as diabetic retinopathy, glaucoma, and age-related macular degeneration. Most affected individuals are over the age of 50, and the associated annual productivity loss is estimated at approximately US\$411 billion \cite{WHO2023Blindness}. Regular screening for diabetic retinopathy is recommended by the World Health Organization and has been shown to be cost-effective across diverse healthcare systems, enabling early detection of sight-threatening complications \cite{AbouTaha2024DRscreening}. Telemedicine-based retinopathy screening programs rely on digital retinal imaging. They have demonstrated cost effectiveness in both high and low resource settings by reducing geographical, financial, and socio economic barriers to care \cite{SACCHINI2025109139}. However, these programs still depend on manual grading by specialists, creating bottlenecks as the number of patients continues to rise. Integrating artificial intelligence (AI) systems for automated screening of retinal images is therefore a key next step to reduce clinician workload, improve efficiency, and further enhance the scalability and cost-effectiveness of screening services \cite{AbouTaha2024DRscreening}.

In recent years, AI has become increasingly important in healthcare, particularly in medical imaging. Deep learning methods based on convolutional neural networks (CNNs) and, more recently, vision transformers (ViTs) have achieved strong performance across a wide range of image classification, segmentation, and computer-aided diagnosis tasks \cite{Chan2020DeepLearningMedImage, Carin2018DeepLearningMedImage}. CNNs learn hierarchical local features from large image collections; ViTs use self-attention to capture global context; hybrid CNN–transformer architectures aim to combine local and global inductive biases; and, more recently, vision–language models (VLMs) such as CLIP leverage image–text pretraining. These families of models have become central to the automated detection of retinal diseases \cite{9783206, Muchuchuti2023RetinalDiseaseReview, Lim2024AIRetinalDiseases, Akil2021DetectionOR}. By providing rapid and reliable image analysis and supporting more efficient care pathways, these tools have important implications for health technology assessment and public health planning. This growing evidence base underscores the need to rigorously benchmark modern architectures-including convolutional, transformer, hybrid, and vision–language models, to understand how they compare in scalable multi-disease retinal screening.

\section{Related work}
In recent years, deep learning has become central to automated diabetic retinopathy screening, with most established systems relying on convolutional neural networks (CNNs) trained on large retinal image datasets \cite{GOH2024100552}. Gulshan and colleagues developed and validated a deep CNN for automated detection of referable diabetic retinopathy and diabetic macular edema from retinal fundus photographs, training on more than 128{,}000 images and evaluating performance on two independent clinical datasets \cite{Gulshan2016JAMA}. Their algorithm achieved an area under the receiver operating characteristic curve (AUC) of 0.991 on the EyePACS-1 validation set and 0.990 on Messidor-2, with sensitivity and specificity exceeding 90\% at operating points optimized for either high specificity or high sensitivity, demonstrating that deep learning can reach specialist-level performance for large-scale diabetic retinopathy screening \cite{Gulshan2016JAMA}. Ting et al. developed and validated a deep learning system for simultaneous detection of referable diabetic retinopathy, vision-threatening diabetic retinopathy, possible glaucoma, and age-related macular degeneration from retinal fundus images in multiethnic populations with diabetes, using nearly half a million images for training and evaluation \cite{Ting2017DevelopmentAV}. In external validation across community and clinic based cohorts, the system achieved AUCs of approximately 0.94 to 0.96 with sensitivities and specificities around or above 90\% for all four target conditions (referable diabetic retinopathy, vision threatening diabetic retinopathy, possible glaucoma, and age related macular degeneration), demonstrating the feasibility of multi disease retinal screening with a single automated model \cite{Ting2017DevelopmentAV}. 

Choi et al. proposed a pilot multi-categorical deep learning framework that used transfer learning–based CNNs with ensemble classifiers to distinguish 10 categories of retinal images, including normal retina and nine retinal diseases on the small STructured Analysis of the REtina (STARE) database, demonstrating both the feasibility and the performance limitations of multi-disease fundus classification under limited data conditions \cite{MultiDeep}. Quellec et al. proposed a framework for automated detection of rare retinal conditions in large-scale diabetic retinopathy screening networks, addressing the limitation that conventional deep learning systems typically focus only on frequent pathologies \cite{QUELLEC2020101660}. Their approach augments CNNs trained on common diseases with an unsupervised probabilistic model over the feature space, leveraging similarities between images to improve recognition of rare findings and enabling reliable detection of 37 out of 41 conditions with an average AUC of 0.938 on more than 160{,}000 screening examinations \cite{QUELLEC2020101660}.

Recent advances in computer vision indicate that architectures beyond traditional CNNs can further improve image classification performance. Dosovitskiy et al. showed that a pure transformer architecture, the Vision Transformer (ViT), can match or surpass state-of-the-art CNNs on benchmarks such as ImageNet, CIFAR-100, and VTAB when pretrained on very large image datasets and then fine-tuned on downstream tasks \cite{dosovitskiy2021an}. In ViT, an image is divided into fixed-size patches that are linearly embedded, combined with positional encodings, and processed by a standard transformer encoder, showing that large-scale training can compensate for the lack of convolutional inductive biases and establishing transformers as a competitive alternative for visual recognition \cite{dosovitskiy2021an}. Vision transformers and hybrid CNN–transformer models have already shown encouraging results in ophthalmic AI. For instance, Goh et al. conducted a comparative analysis of multiple state-of-the-art CNNs and ViTs on 48{,}269 retinal photographs from the Kaggle diabetic retinopathy dataset, Messidor-1, and the Singapore Epidemiology of Eye Diseases study. They found that transformer-based architectures-particularly the Swin Transformer-consistently achieved higher AUC and sensitivity than CNNs for detecting referable diabetic retinopathy across both internal and external test sets \cite{GOH2024100552}.

In addition to vision transformers, vision-language models (VLMs) such as CLIP (Contrastive Language–Image Pretraining) have gained prominence for learning transferable visual representations from natural language supervision \cite{Radford2021LearningTV, 10445007}. CLIP is trained contrastively on a dataset of roughly 400 million image–text pairs collected from the internet, aligning images and captions in a shared embedding space and enabling strong zero-shot performance across a wide range of image classification benchmarks without task-specific fine-tuning \cite{Radford2021LearningTV}. Although originally developed for natural images, its image encoder can be repurposed as a generic feature extractor, making such models attractive candidates for improving data efficiency and out-of-distribution generalization in medical imaging applications, including retinal disease screening.

Despite this rapid progress, several important gaps remain in the retinal imaging literature. Most prior studies either focus on single disease screening tasks or evaluate only CNN or transformer-based architectures in isolation. When comparative analyses are performed, they are usually restricted to diabetic retinopathy endpoints such as referable diabetic retinopathy or vision-threatening diabetic retinopathy \cite{GOH2024100552, Gulshan2016JAMA, Ting2017DevelopmentAV, MultiDeep, QUELLEC2020101660}.
Even when multiple architectures are compared, they are usually conventional CNNs or vision transformers trained in a supervised fashion, without considering hybrid CNN–transformer backbones or large-scale VLMs trained on natural image–text pairs. Although VLMs have emerged as powerful, generic image encoders \cite{Radford2021LearningTV} and hybrid CNN–transformer architectures have begun to be explored for retinal disease classification \cite{Lisha2025HybridRetinal}, their suitability for clinically realistic multi-disease retinal screening and their relative performance against strong CNN and transformer baselines on datasets like RFMiD remain largely unexplored. In particular, there is limited evidence on whether hybrid CNN-transformer models and VLMs improve detection performance in multi-disease retinal screening, enhance robustness to dataset shifts, or offer better data efficiency than traditional architectures. This gap motivates our systematic, head-to-head comparison of CNNs, ViTs, hybrid models, and VLMs on the Retinal Fundus Multi-disease Image Dataset (RFMiD) for both a binary screening task and a multi-label classification task.

The RFMiD was introduced by Pachade et al. to address the lack of publicly available datasets that capture both frequent and rare ocular pathologies encountered in routine clinical practice \cite{RFMiD}. RFMiD consists of 3{,}200 color fundus photographs acquired with three different fundus cameras, covering 46 distinct conditions annotated through adjudicated consensus of two senior retinal experts \cite{RFMiD}. It is specifically designed to support both a binary screening task that distinguishes normal from abnormal images and a multi-label classification task that detects a broad range of retinal diseases, thereby enabling the development and benchmarking of generalizable AI models for comprehensive retinal screening \cite{RFMiD}.

In this work, we leverage the RFMiD dataset to perform a systematic comparison of multiple deep learning architectures for retinal pathology classification. We evaluate four CNN models (ResNet50 \cite{7780459}, InceptionV3 \cite{inproceedings}, DenseNet121 \cite{8099726}, and EfficientNetB3 \cite{pmlr-v97-tan19a}); four transformer-based models (CrossViT-Small \cite{crossViTSmall}, DeiT-Small \cite{DeitSmall}, Swin-Tiny \cite{swin}, and ViT-Small \cite{vitSmall}); two hybrid models (CoAtNet-0 \cite{coatnet0} and MaxViT-Tiny \cite{MaxViT}); and two vision-language models (CLIP ViT-B/16 \cite{Radford2021LearningTV} and SigLIP Base-384 \cite{siglip}) by fine-tuning all models on the same RFMiD training set. The primary objective is to assess performance on a binary screening task, which determines whether a fundus image contains any retinal disease and therefore would trigger further clinical review. The secondary objective is to evaluate model performance on a multi-label classification task, in which each model predicts the full spectrum of RFMiD pathologies present in an image. By providing a head-to-head comparison of CNNs, ViTs, hybrid CNN–transformer backbones, and VLMs on a clinically relevant multi-disease retinal dataset, this study aims to clarify the respective strengths and limitations of each architecture family. In doing so, it offers practical guidance for selecting model backbones and designing robust, generalizable retinal screening systems that can support both large-scale triage and more detailed disease-wise interpretation in real-world settings.

\section{Methodology}

\begin{figure*}[t]
  \centering
  \includegraphics[width=\textwidth]{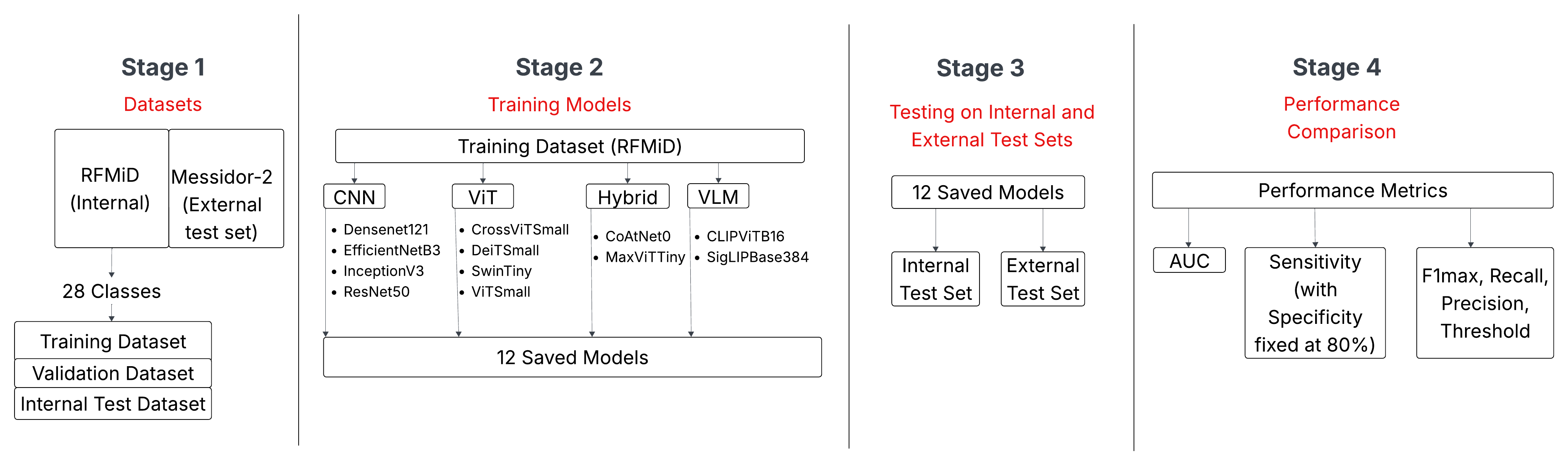}
  \caption{Overview of the experimental methodology. Stage 1: datasets (RFMiD for internal training/validation/testing and Messidor-2 as external test set). Stage 2: training of 12 CNN, ViT, hybrid, and VLM models on the RFMiD training set. Stage 3: evaluation on the RFMiD internal test set and the Messidor-2 external test set. Stage 4: performance comparison using AUC and sensitivity at fixed specificity, together with F1, precision, recall, and decision thresholds.}
  \label{fig:overview}
\end{figure*}

We begin with data preparation, including dataset selection and preprocessing. We then fine-tune multiple pretrained convolutional, hybrid CNN transformer, pure transformer, and vision language architectures. We optimize model parameters on the training split and use the validation split to guide model selection and operating point calibration.
Finally, we evaluate the selected models on the held-out RFMiD test set. In addition, we assess out-of-domain generalization by evaluating all RFMiD-trained models on the Messidor-2 dataset \cite{Decenciere2014Messidor}. This unified pipeline ensures that all model families are assessed under identical data splits, calibration procedures, and metrics, enabling a fair head-to-head comparison. The following sections detail the procedures, settings, and evaluation criteria. An overview of the experimental pipeline-including datasets, model families, evaluation on internal and external test sets, and performance metrics-is shown in Figure~\ref{fig:overview}.

\subsection{Dataset and preprocessing}

We use the Retinal Fundus Multi-disease Image Dataset (RFMiD) \cite{RFMiD}, which provides official training, validation, and test splits with per-image multi-label annotations for retinal pathologies. The set of disease classes is defined by all columns except \texttt{ID} in the label CSV files. We use the three label files exactly as released: a.RFMiD\_Training\_Labels.csv, b.RFMiD\_Validation\_Labels.csv, and c.RFMiD\_Testing\_Labels.csv. Images are read from the corresponding split folders under \texttt{Original Images} and converted to three-channel RGB using the Python Imaging Library (PIL) with antialiasing. Model-specific resizing and normalization are applied by the training and evaluation transforms. We consider two main prediction tasks: (i) a binary screening task and (ii) a multi-label classification task. For completeness, we report both overall any abnormal metrics and per-class metrics, and we use the validation split to select operating points.

\subsubsection{Dataset}

The RFMiD dataset used in this work is a public collection of 3{,}200 color fundus photographs with per-image, multi-label annotations for retinal pathologies. Images were acquired using three fundus cameras: TOPCON 3D OCT 2000 ($45^{\circ}$, $2144 \times 1424$), Kowa VX 10$\alpha$ ($50^{\circ}$, $4288 \times 2848$), and TOPCON TRC NW300 ($45^{\circ}$, $2048 \times 1536$). The dataset intentionally includes both high- and low-quality images and reflects the variability encountered in routine clinical practice.

RFMiD is released in two configurations: an \textit{all-classes} version with 46 annotated conditions, and a \textit{challenge} subset tailored for benchmarking. In this work, we use the RFMiD challenge subset. In the challenge subset, the images are split into training (1{,}920), validation (640), and test (640) sets. Disease categories with more than 10 images are retained as independent classes, while the remaining rare conditions are merged into a single \textsc{Other} category, yielding 28 classes in total \cite{RFMiD}. The 28 challenge classes span a broad spectrum of retinal diseases, including common entities such as diabetic retinopathy (DR), age-related macular degeneration (ARMD), myopia (MYA), branch retinal vein occlusion (BRVO), and tessellation (TSLN), as well as rarer but sight-threatening conditions such as central retinal artery occlusion (CRAO) and optic disc pit maculopathy (ODPM) \cite{RFMiD}.

To assess out-of-domain generalization, we additionally evaluate all RFMiD-trained models on the Messidor-2 dataset \cite{Decenciere2014Messidor} using a preprocessed version obtained from Kaggle. Messidor-2 consists of macula-centered color fundus photographs with adjudicated DR grades and gradability flags, and is widely used as an external benchmark for diabetic retinopathy screening \cite{Gulshan2016JAMA, Ting2017DevelopmentAV}. In our experiments, we follow the Messidor-2 grading scheme and focus on a referable DR endpoint (diagnosis $\geq 2$) for external validation. We chose Messidor-2 as our external test set for two main reasons: (i) it is one of the most established public benchmarks in the diabetic retinopathy literature, and its expert-adjudicated referable DR labels provide a clinically meaningful endpoint that aligns with the binary screening scenario studied on RFMiD; and (ii) using a single-disease dataset with different acquisition protocols and prevalence helps us highlight the effect of domain and label set shift, and provides a controlled setting to compare how CNN, transformer, hybrid, and vision–language architectures maintain or lose performance under distribution shift.

\subsubsection{Data processing}

We formulate the RFMiD challenge subset as a multi-label classification problem over 28 classes, and we also define a binary screening task that distinguishes normal from abnormal images. We follow the official training, validation, and test CSV files provided with the dataset. We first use the training label file to define the list of disease labels and their column order. We then reorder the validation and test label files to match this schema, adding any missing columns and filling missing entries with zero, which we treat as the absence of the corresponding pathology.

Images are loaded as three-channel RGB images by identifier. For the hybrid backbones CoAtNet0 and MaxViTTiny, we use the model-specific pretrained configurations from the PyTorch Image Models (\texttt{timm}) library: random resized crop to 224 pixels, horizontal flip, and normalization during training; and resize plus center crop with the same normalization during evaluation. For CLIP ViT-B/16 and SigLIP Base-384, we use their native transforms with the corresponding input size and normalization, applying the same light augmentation during training. CNN and ViT baselines use analogous \texttt{timm} transforms at their respective input resolutions.

Labels are stored as a binary vector per image. We address class imbalance using per-class positive weights computed from the training split and optimize a binary cross-entropy with logits loss. Training uses mixed precision when available, gradient clipping at norm 1.0, and a fixed seed for reproducible data order and weight initialization.

After training, we perform temperature scaling on the validation logits and compute per-class decision thresholds targeting a validation specificity of approximately 0.80. The full calibration procedure, including both the any-abnormal versus normal binary screening task and the multi-label, disease-wise task, is described in the evaluation section.

\subsection{Model selection and architecture}

We compare four families of vision models on RFMiD: convolutional neural networks (CNNs), hybrid CNN--transformer backbones, Vision Transformer (ViT) architectures, and vision-language models (VLMs). Each family brings a distinct inductive bias and feature representation strategy, allowing us to test whether local texture cues, global context, or language-aligned embeddings better capture retinal pathology. Our model selection was guided by both clinical relevance and architectural diversity. We include widely adopted and well-supported models from each family that represent state-of-the-art or influential baselines in their respective categories (e.g., ResNet50, SwinTiny, CLIP ViT-B/16). Rather than exhaustively covering all possible models, we focus on representative architectures that are feasible to deploy and fine-tune using standard open-source tools. This approach enables a fair and interpretable comparison across design paradigms, while ensuring that findings are relevant for real-world retinal screening applications.

\textbf{CNN models.} As classical baselines, we include ResNet50 \cite{7780459}, InceptionV3 \cite{inproceedings}, DenseNet121 \cite{8099726}, and EfficientNetB3 \cite{pmlr-v97-tan19a}. These models provide strong performance on natural images with well-understood training behavior and serve as a reference point for newer architectures. Their hierarchical receptive fields are well-suited to detecting vessel caliber, exudates, hemorrhages, and other local retinal features.

\textbf{Hybrid models.} We evaluate CoAtNet0 \cite{coatnet0} and MaxViTTiny \cite{MaxViT}, which combine convolutional stages for low-level pattern extraction with attention blocks for longer-range interactions. These hybrids aim to retain translation-friendly local processing while enabling global context aggregation across the fundus. In our implementation, the backbone outputs a single feature vector per image, which is passed to a lightweight classifier with dropout and ReLU, trained using binary cross-entropy with logits across all classes.

\textbf{ViT models.} We include four transformer-based vision architectures to isolate the contribution of self-attention: CrossViT-Small \cite{crossViTSmall}, DeiT-Small \cite{DeitSmall}, Swin-Tiny \cite{swin}, and ViT-Small \cite{vitSmall}. These models use patch embeddings and multi-head self-attention to capture long-range structure, which is particularly relevant for modeling macula-to-disc relationships and diffuse abnormalities.

\textbf{VLM models.} We examine CLIP ViT-B/16 \cite{Radford2021LearningTV} and SigLIP-Base384 \cite{siglip} as representatives of image–text pretraining. For CLIP, we keep the text encoder frozen and form per-class text embeddings from short retinal prompts. Image logits are computed as cosine similarities between normalized image and text embeddings, scaled by a learned temperature and adjusted by a small per-class bias. For SigLIP, we use the vision backbone with a small classifier head. These models test whether language-aligned features transfer effectively to fundus classification with limited task-specific tuning.

\textbf{Training heads and selection protocol.} All models use a consistent per-image multi-label classifier head and are optimized with binary cross-entropy with logits. For CLIP, we follow a two-stage training schedule: we first train only the learned logit scale and per-class bias, then unfreeze the final transformer blocks of the vision encoder and continue training them with a reduced learning rate. For all other backbones, we train the classifier head with a higher learning rate and the backbone with a lower learning rate. Model selection is based on validation area under the ROC curve, followed by temperature scaling on the validation logits and per-class threshold calibration targeting a specificity of approximately 0.80. These selected operating points are then applied to the held-out test set.

\subsection{Model training and optimization}

All models are trained on the official RFMiD training split, with the validation split used for model selection and operating point calibration. We adopt a shared optimization framework across model families, with family-specific modifications as needed, and follow standard training practices from \texttt{timm} and the original model repositories.

\textbf{Optimization and schedules.}
We use AdamW with a cosine learning rate schedule and a short warmup period.
For CNN, ViT, and hybrid backbones, the classifier head is trained with a higher learning rate than the feature extractor.
Unless otherwise specified, the backbone uses a learning rate of \(5 \times 10^{-5}\) with weight decay of 0.05, and the classifier head uses \(5 \times 10^{-4}\) with no weight decay.
Vision–language models follow a two-stage training procedure.
For CLIP, we first train only the logit scaling parameter and a small per-class bias, then unfreeze the final transformer blocks in the vision encoder and continue fine-tuning them at a lower learning rate.
For SigLIP, we initially train the small classifier head and optionally unfreeze the final vision layers with a reduced learning rate.
Cosine learning rate schedules with warmup are applied in all cases.

\textbf{Loss and imbalance handling.}
We formulate RFMiD as a multi-label classification problem and optimize binary cross-entropy with logits.
To address class imbalance, we compute per-class positive weights from the training data and incorporate them into the loss function.
Labels are stored as binary vectors per image, aligned with the schema defined by the training label file, and are consistently applied to the validation and test sets.

\textbf{Data preprocessing and augmentation.}
Images are read as three-channel RGB PNGs by identifier.
For hybrid backbones, we apply random resized crops to \(224 \times 224\) pixels, horizontal flipping, and model-specific normalization during training, and use resizing with center cropping and the same normalization during evaluation.
CLIP ViT-B/16 and SigLIP Base384 use their native preprocessing pipelines, including input size and normalization, with light data augmentation during training.
CNN and ViT baselines adopt \texttt{timm}-standard transforms appropriate to their respective input resolutions.

\textbf{Stability and regularization.}
We use mixed-precision training when available, gradient clipping with a maximum norm of 1.0, and fixed seeds for reproducible data shuffling and weight initialization.
The batch size is 16 for CNN, ViT, and hybrid models.
For vision–language models, we use a batch size of 4 with gradient accumulation over 4 steps to achieve an effective batch size of 16 while managing GPU memory constraints.
Early stopping is based on validation AUROC with a patience of 10 epochs and a minimum improvement threshold of \(1 \times 10^{-4}\).
To reduce overfitting, we cap training at 20 epochs, monitor divergence between training and validation loss, and apply regularization via data augmentation, dropout in the classifier head, and AdamW weight decay.

\textbf{Calibration and threshold selection.}
After training, we perform temperature scaling on the validation logits using a limited-memory BFGS  (L-BFGS) optimizer to learn a single scalar temperature per model.
We then compute per-class decision thresholds by scanning the validation ROC curves and selecting the point where specificity is closest to 0.80.
These calibrated temperature values and per-class thresholds are fixed and directly applied to the held-out test set.

\textbf{Selection criterion.}
During training, we monitor the AUROC on the validation split at the end of each epoch.
Whenever the validation AUROC improves, we save the corresponding model checkpoint.
At the end of training, we select the checkpoint with the highest validation AUROC, and all results reported in the Results section are computed using this selected model.

\subsection{Evaluation metrics}

We report metrics for two tasks: a binary screening task and a multi-label classification task. In the binary task, a fundus image is assigned a single binary label: abnormal if at least one RFMiD disease label is present, and normal otherwise. In the multi-label setting, the model predicts a separate binary label for each RFMiD disease category, allowing multiple coexisting pathologies per image. Unless otherwise stated, calibration and threshold selection are performed on the validation split and applied unchanged to the held-out test set.

\textbf{Calibration.}
For each model we apply temperature scaling to the validation logits,
learning a single scalar temperature $T$ with L-BFGS to minimize the
binary cross-entropy loss. At test time, logits are divided by $T$
before computing any metrics.

\textbf{ROC-based metrics.}
For each disease label $c$ we compute the area under the ROC curve (AUC).
We report a macro AUC as the mean over all labels that have both
positive and negative samples,
\[
\mathrm{AUC}_{\text{macro}}
= \frac{1}{|C'|} \sum_{c \in C'} \mathrm{AUC}_c,
\]
where $C'$ is the set of such labels and $\mathrm{AUC}_c$ is the AUC
for label $c$.
We also report a micro AUC by pooling all per-label predictions into a
single contingency table. Concretely, we form the pooled set
\[
\mathcal{D}_{\text{micro}}
= \big\{(y_{ic}, \hat{p}_{ic}) : 1 \leq i \leq N,\; c \in C' \big\},
\]
where $y_{ic} \in \{0,1\}$ is the ground-truth label and $\hat{p}_{ic}$
is the calibrated probability, and compute
\[
\mathrm{AUC}_{\text{micro}} = \mathrm{AUC}(\mathcal{D}_{\text{micro}}),
\]
that is, the standard AUC computed on all pairs $(y_{ic}, \hat{p}_{ic})$
concatenated across labels.

For the binary screening task, we form a single
label per image by taking the logical OR over all disease labels,
\[
y_i^{\text{any}} = \mathbf{1}\Big( \bigvee_{c} y_{ic} = 1 \Big),
\]
and define the image score as the maximum calibrated class probability
$s_i^{\text{any}} = \max_{c} \hat{p}_{ic}$.
We then report AUC on this binary task from the ROC curve obtained by
sweeping a threshold over $s_i^{\text{any}}$ \cite{Fawcett2006ROC}.
Here, $y_{ic} \in \{0,1\}$ denotes the ground truth label indicating whether
image $i$ has disease $c$, $\hat{p}_{ic}$ is the corresponding calibrated
probability, and $\mathbf{1}(\cdot)$ is the indicator function.

\textbf{Operating point selection.}
For the binary screening task, we select a
single decision threshold on the validation set by sweeping the
threshold over $[0,1]$, computing precision and recall for each value,
and choosing the threshold that maximizes the F1 score. Given true
positives $\mathrm{TP}(t)$, false positives $\mathrm{FP}(t)$ and false
negatives $\mathrm{FN}(t)$ at threshold $t$, precision, recall and F1 are
\[
\mathrm{Prec}(t) = \frac{\mathrm{TP}(t)}{\mathrm{TP}(t) + \mathrm{FP}(t)}, \qquad
\mathrm{Rec}(t) = \frac{\mathrm{TP}(t)}{\mathrm{TP}(t) + \mathrm{FN}(t)},
\]
\[
F1(t) = \frac{2\,\mathrm{Prec}(t)\,\mathrm{Rec}(t)}{\mathrm{Prec}(t) + \mathrm{Rec}(t)}.
\]
We then define the validation-selected decision threshold as
\[
\tau_{\text{any}} = \arg\max_{t \in [0,1]} F1(t),
\]
and keep this value fixed when evaluating on the test set.

For the multi-label classification task, we instead calibrate a separate
decision threshold $\tau_c$ for each disease label by scanning the
validation ROC curve to find the operating point whose specificity is
closest to $0.80$. For label $c$, with true negatives
$\mathrm{TN}_c(t)$ and false positives $\mathrm{FP}_c(t)$ at threshold $t$,
specificity is
\[
\mathrm{Spec}_c(t)
= \frac{\mathrm{TN}_c(t)}{\mathrm{TN}_c(t) + \mathrm{FP}_c(t)},
\]
and we choose
\[
\tau_c = \arg\min_{t \in [0,1]} \big|\mathrm{Spec}_c(t) - 0.80\big|.
\]
These per class thresholds $\{\tau_c\}$ are reused unchanged on the test set \cite{Powers2011Eval}.

\textbf{Threshold-based metrics.}
Given calibrated probabilities $\hat{p}_{ic}$ and thresholds, we
compute per-class sensitivity and specificity, along with their macro
averages, and micro sensitivity and specificity by pooling counts
across all labels. For a given class $c$ at threshold $t$, with true
positives $\mathrm{TP}_c(t)$, false negatives $\mathrm{FN}_c(t)$,
true negatives $\mathrm{TN}_c(t)$, and false positives
$\mathrm{FP}_c(t)$, sensitivity, specificity, and balanced accuracy are
\[
\mathrm{Sens}_c(t)
= \frac{\mathrm{TP}_c(t)}{\mathrm{TP}_c(t) + \mathrm{FN}_c(t)}, \quad
\mathrm{Spec}_c(t)
= \frac{\mathrm{TN}_c(t)}{\mathrm{TN}_c(t) + \mathrm{FP}_c(t)},
\]
\[
\mathrm{BA}_c(t)
= \tfrac{1}{2}\big(\mathrm{Sens}_c(t) + \mathrm{Spec}_c(t)\big).
\]
From the resulting confusion matrix, we also derive overall TP, TN, FP,
and FN. For the binary screening task, we
report precision, recall, F1, and confusion matrix entries at
$\tau_{\text{any}}$, together with $F1_{\max}$ and its associated
threshold. For the multi-label classification task, we report macro and
micro F1 on the test set using the per-class thresholds $\{\tau_c\}$
calibrated at approximately 0.80 specificity \cite{Powers2011Eval}.

\textbf{Reporting.}
In the Results section, binary performance for each model is summarized by the AUC for the binary screening task, $F1_{\max}$, and the corresponding precision, recall, and decision threshold.
Multi-label classification performance is
summarized by macro and micro AUC (threshold-free) and macro and
micro F1 computed at the specificity calibrated thresholds. In
addition, we report sensitivity at a fixed specificity of
approximately 80\% to reflect a clinically relevant operating point
and use grouped bar plots to compare AUC and sensitivity across model
families. External validation on Messidor-2 is reported using the
same set of metrics to enable a direct comparison of performance
under domain and label set shift.

\subsection{Implementation and code availability}

All models are implemented in PyTorch \cite{paszke2019pytorch}. CNN, ViT, and hybrid backbones
use the \texttt{timm} library for architecture definitions and
pretrained weights, while CLIP ViT/B16 and SigLIP Base-384 follow
their official open-source implementations.
Training and
evaluations are carried out with a unified pipeline that handles data
loading, augmentation, loss weighting, optimization schedules,
temperature scaling, and threshold calibration in a consistent way
across all model families. To facilitate reproducibility and future research, we release our
training and evaluation code, together with configuration files
specifying all training settings (learning rates, batch sizes, number
of epochs, data augmentations, and optimizer parameters) and scripts
for RFMiD and Messidor-2 preprocessing, at:
\url{https://github.com/Durjoy001/Retinal-NeuralNET}.
 The repository
also includes example command lines to reproduce the main experiments
and generate the tables and figures reported in this paper.

\section{Results}

\begin{table*}[t]
\centering
\setlength{\tabcolsep}{3.5pt} % tighter columns but normal font size
\caption{RFMiD test performance of all visual architectures.
Metrics for the binary screening task are reported at the $F1_{\max}$ operating point selected on the validation set.
For the multi-label classification task, AUC$_\text{macro}$ and AUC$_\text{micro}$ are computed from ROC curves, while $F1_\text{macro}$ and $F1_\text{micro}$ use per-class thresholds calibrated on the validation set to achieve approximately 80\% specificity.
\textbf{Bold} values indicate the best result for each column.}

\label{tab:all_models_results}

\begin{tabular}{ll|ccccc|cccc}
\toprule
\multicolumn{2}{c|}{Model}
  & \multicolumn{5}{c|}{Binary screening task (at $F1_{\max}$)} 
  & \multicolumn{4}{c}{Multi-label classification} \\
\cmidrule(lr){1-2} \cmidrule(lr){3-7} \cmidrule(lr){8-11}
Family & Architecture
  & AUC (\%)
  & F1$_\text{max}$
  & Recall (\%)
  & Precision (\%)
  & Threshold
  & AUC$_\text{macro}$ (\%)
  & AUC$_\text{micro}$ (\%)
  & F1$_\text{macro}$ (\%)
  & F1$_\text{micro}$ (\%) \\
\midrule
CNN    & ResNet50          & 95.4 & 0.95 & 94.9 & 94.5 & 0.58 & 86.2 & 91.8 & 24.6 & 45.3 \\
CNN    & DenseNet121       & 89.4 & 0.91 & 93.1 & 88.9 & 0.68 & 84.5 & 92.7 & 22.8 & 45.8 \\
CNN    & EfficientNetB3    & 84.4 & 0.89 & \textbf{99.8} & 79.9 & 0.55 & 79.4 & 90.3 & 21.0 & 41.8 \\
CNN    & InceptionV3       & 85.8 & 0.89 & 98.0 & 81.9 & 0.53 & 81.0 & 89.2 & 21.5 & 36.7 \\
\midrule
ViT    & CrossViTSmall     & 95.2 & 0.94 & 93.9 & 93.5 & 0.66 & 91.1 & 96.3 & 27.6 & 45.7 \\
ViT    & DeiTSmall         & 96.6 & 0.95 & 97.8 & 91.7 & 0.72 & 90.6 & 96.4 & 30.2 & 47.3 \\
ViT    & SwinTiny          & \textbf{97.8} & \textbf{0.96} & 95.9 & \textbf{97.0} & \textbf{0.90} & \textbf{91.1} & \textbf{97.0} & 30.5 & \textbf{50.9} \\
ViT    & ViTSmall          & 95.2 & 0.94 & 97.6 & 90.2 & 0.59 & 89.4 & 95.0 & 29.4 & 46.4 \\
\midrule
Hybrid & CoAtNet0          & 96.6 & 0.95 & 95.5 & 94.0 & 0.67 & 89.7 & 95.0 & \textbf{34.0} & 50.0 \\
Hybrid & MaxViTTiny        & 96.6 & 0.95 & 94.5 & 96.2 & 0.71 & 90.0 & 95.0 & 30.3 & 47.5 \\
\midrule
VLM    & CLIP ViTB16       & 94.6 & 0.93 & 94.9 & 92.0 & 0.56 & 88.4 & 93.2 & 25.1 & 41.9 \\
VLM    & SigLIPBase384     & 95.5 & 0.94 & 95.3 & 93.2 & 0.65 & 86.9 & 93.5 & 25.4 & 45.8 \\
\bottomrule
\end{tabular}
\end{table*}

In this section, we report the performance of twelve visual model architectures for automated classification of retinal pathologies on the RFMiD challenge dataset. We first describe the evaluation protocol used across all experiments (Section 4.1). We then present results for the binary screening task (Section 4.2) and the multi-label classification task (Section 4.3). Next, we compare the performance of convolutional networks, vision transformers, hybrid CNN–transformer models, and vision–language models, and we summarize the overall ranking of architectures (Section 4.4). Finally, we assess external generalization using the Messidor-2 benchmark (Section 4.5).

\subsection{Evaluation protocol}

All experiments are conducted on the RFMiD retinal fundus image dataset, which provides predefined training, validation, and test splits. The training set is used to fit model parameters, while the validation set is used for early stopping, temperature scaling, and threshold calibration. All numerical results reported in this section are computed on the held-out test set.

We consider two prediction tasks: (i) a binary screening task, and (ii) a multi-label classification task. In the binary screening task, the goal is to determine whether a fundus image is normal or exhibits any abnormal retinal pathology. In the multi-label classification task, the model predicts the presence or absence of each RFMiD disease category per image, allowing multiple pathologies to be labeled simultaneously.

Calibration, operating point selection, and evaluation metrics-including AUC, macro and micro AUC, $F1_{\max}$, macro and micro F1, and sensitivity and specificity-follow the definitions introduced in Section~3.4 \emph{Evaluation Metrics} and are applied identically across all model families.

All models are trained using the same data splits and optimization settings within each architecture family. Their performance is evaluated using a unified protocol and consistent thresholds. This ensures a fair, clinically meaningful comparison between convolutional networks, vision transformers, hybrid CNN–transformer models, and vision-language models.

\subsection{Binary screening task performance}

Table~\ref{tab:all_models_results} summarizes test performance for the binary screening task, which determines whether a fundus image contains any retinal disease and therefore warrants further clinical review. All twelve architectures achieve strong discrimination, with AUC values above 84\%, though clear differences emerge between model families. Among the CNN baselines, ResNet50 performs best (AUC 95.4\%, $F1_{\max}=0.95$), whereas EfficientNetB3 and InceptionV3 trail behind with AUCs of 84.4\% and 85.8\%, respectively. Vision transformer and hybrid models consistently perform better: SwinTiny achieves the highest AUC (97.8\%) and the highest $F1_{\max}$ (0.96), followed closely by DeiTSmall, CoAtNet0, and MaxViTTiny (AUC $\approx$ 96.6\%). Vision–language models are competitive but slightly behind the top ViT and hybrid models (CLIP ViT-B/16: AUC 94.6\%, SigLIP-Base384: AUC 95.5\%).

At the $F1_{\max}$ operating point, most models achieve a favorable balance between recall and precision. EfficientNetB3 yields the highest recall (99.8\%) but with lower precision (79.9\%), suggesting a more aggressive decision threshold that results in more false positives. In contrast, SwinTiny combines high recall (95.9\%) with the highest precision (97.0\%), resulting in the best overall $F1_{\max}$. The hybrid models CoAtNet0 and MaxViTTiny also provide similarly balanced and robust performance.

\begin{figure*}[t]
  \centering
  \includegraphics[width=\textwidth]{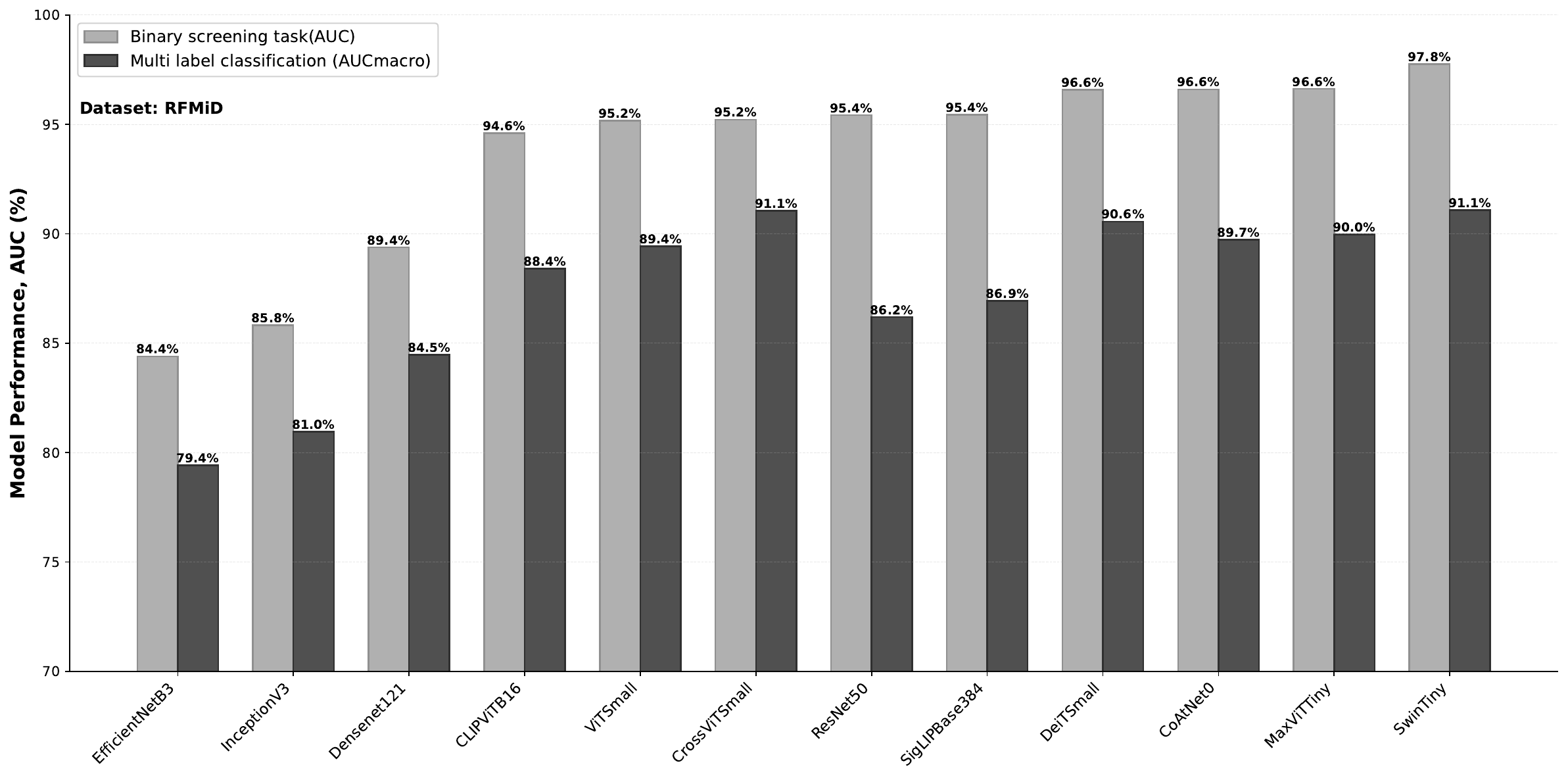}
  \caption{Comparison of AUC for the binary screening task and the multi-label classification task across all 12 models on the RFMiD dataset.}
    \Description{Grouped bar chart comparing 12 deep learning models for retinal image analysis. 
    For each model, the left bar shows AUC for any abnormal versus normal detection, and the right bar shows macro AUC for multi label disease wise classification. 
    SwinTiny has the highest AUC in both tasks, followed by other ViT and hybrid models, while some CNNs achieve lower AUC values. 
    The x-axis lists model names and the y-axis shows AUC in percent from 70 to 100.}
  \label{fig:model_auc_comparison}
\end{figure*}

\begin{figure*}[t]
  \centering
  \includegraphics[width=\textwidth]{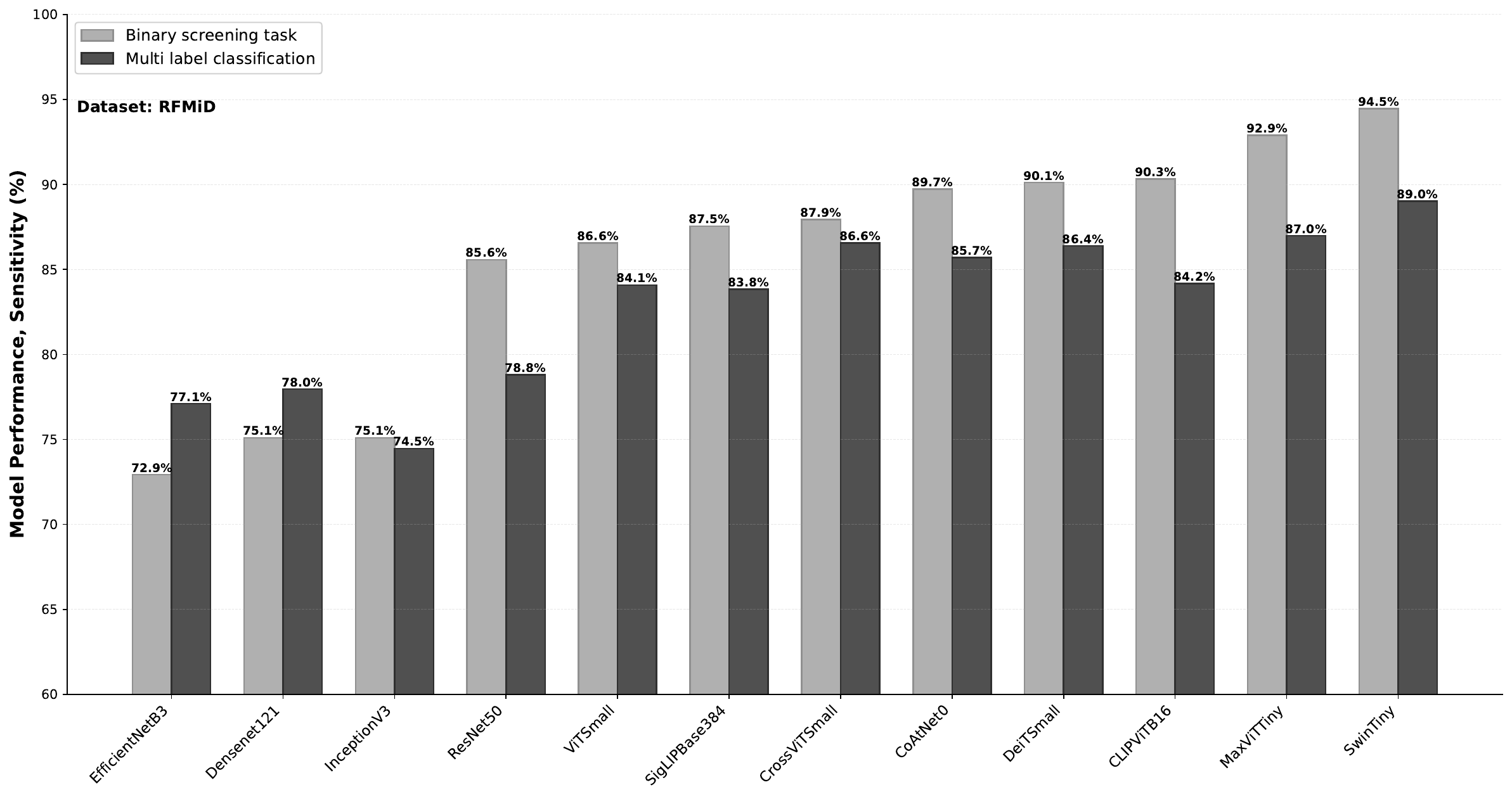}
\caption{Comparison of model sensitivity (with specificity fixed at 80\%) for the binary screening task and the multi-label classification task across all 12 models on the RFMiD dataset.}
  \Description{Grouped bar chart showing, for each of the 12 models, two bars:
  one for sensitivity on the any abnormal versus normal binary screening task
  and one for sensitivity on the multi label disease wise classification task,
  both evaluated at approximately 80 percent specificity. Transformer and hybrid
  models generally achieve higher sensitivity than CNN and vision–language baselines.}
  \label{fig:model_sensitivity_comparison}
\end{figure*}

Figure~\ref{fig:model_auc_comparison} visualizes the AUC values for all architectures, highlighting the consistent advantage of transformer and hybrid backbones over classical CNNs and VLMs for this binary screening task. When fixing the operating point to a clinically relevant specificity of approximately 80\% using validation-calibrated thresholds, the relative ranking remains largely unchanged. As shown in Figure~\ref{fig:model_sensitivity_comparison}, SwinTiny, CoAtNet0, and MaxViTTiny maintain the highest sensitivities, while CNN baselines show lower sensitivity at the same specificity.

Together, these results suggest that attention-based architectures-especially ViTs and hybrids-are better suited than purely convolutional networks for detecting any abnormal retinal pathology in the RFMiD dataset.

\subsection{Multi-label classification performance}

For the multi-label classification setting, each model outputs a separate binary prediction for every RFMiD disease category in each image. This setup allows a single fundus photograph to be assigned multiple concurrent pathologies, reflecting the real-world clinical scenario where patients may present with several retinal conditions simultaneously. As reported in Table~\ref{tab:all_models_results}, this task is substantially more challenging than binary screening: macro F1 scores range from approximately 21\% to 34\%, reflecting the strong class imbalance and the presence of rare pathologies—even though ranking performance remains high (macro AUC $>79\%$ for all models).

Among the CNN baselines, ResNet50 again performs best, with a macro AUC of 86.2\% and a macro F1 of 24.6\%, slightly outperforming DenseNet121, EfficientNetB3, and InceptionV3. Vision transformers and hybrid backbones provide a clear boost in multi-label classification. SwinTiny achieves the highest macro and micro AUC (91.1\% and 97.0\%, respectively) and the highest micro F1 (50.9\%), indicating strong performance on common and moderately prevalent diseases. CoAtNet0 attains the best macro F1 (34.0\%) alongside competitive AUC values (89.7\% macro, 95.0\% micro), suggesting improved handling of rare pathologies when all classes are weighted equally. DeiTSmall, CrossViTSmall, and MaxViTTiny also outperform all CNNs across both AUC and F1 metrics.

The vision–language models yield intermediate performance. CLIP ViT-B/16 and SigLIP-Base384 achieve macro AUCs of 88.4\% and 86.9\% and macro F1 scores around 25\%, performing comparably to the stronger CNN baselines but consistently below the top transformer and hybrid models. This pattern is mirrored in the sensitivity results at fixed 80\% specificity in Figure~\ref{fig:model_sensitivity_comparison}, where transformer and hybrid architectures maintain higher sensitivity than CNN and VLM baselines.

Overall, these findings indicate that attention-based backbones-particularly SwinTiny and CoAtNet0-are better suited for capturing the diverse spectrum of retinal pathologies in RFMiD.

\subsection{Overall comparison of model families (CNN, ViT, Hybrid, VLM)}

Table~\ref{tab:all_models_results} and Figures~\ref{fig:model_auc_comparison}--\ref{fig:model_sensitivity_comparison} summarize the overall performance of the four model families across both evaluation tasks. All architectures achieve reasonable results on the binary screening task, with AUC values above 84\% for every model. Within this task, classical CNNs provide a strong baseline (e.g., ResNet50, AUC 95.4\%), but attention-based models consistently outperform them. SwinTiny achieves the highest binary AUC (97.8\%) and $F1_{\max}$ (0.96), while CoAtNet0 and MaxViTTiny obtain similarly strong AUC and F1 scores, suggesting that transformer blocks offer consistent gains even over carefully tuned CNNs.

Performance differences become more pronounced in the multi-label classification task. While CNNs show a clear drop in macro F1 (around 21--25\%), hybrid and transformer architectures maintain substantially higher scores. CoAtNet0 achieves the highest macro F1 (34.0\%), and SwinTiny leads in micro F1 (50.9\%) along with the best macro and micro AUC, indicating strong performance across both frequent and rare disease classes. Vision–language models (CLIP ViT-B/16 and SigLIP-Base384) perform between CNNs and the best transformer or hybrid models: they modestly outperform weaker CNNs but do not match the top-performing ViTs or hybrids.

Taken together, these results suggest a consistent performance hierarchy across tasks: transformer and hybrid CNN–transformer models provide the strongest results, CNNs offer competitive yet weaker baselines, and current vision-language models, when adapted with straightforward classification heads, do not yet match specialized vision transformers on RFMiD. For practical deployment in retinal screening systems, Swin-style or hybrid backbones are preferable when computational resources permit, while ResNet50 remains a reasonable option for resource-constrained settings.

\subsection{External dataset generalization}

\begin{table*}[t]
\centering
\setlength{\tabcolsep}{4pt} % tighter columns but normal font size
\caption{(External dataset performance) Referable diabetic retinopathy (referable DR; Messidor-2 diagnosis $\geq 2$) detection performance of all visual model
architectures on the Messidor-2 dataset. All models are trained on RFMiD and
evaluated on Messidor-2 without further weight updates. AUC is computed on the
gradable test set for the referable DR endpoint. For each model, the threshold $\tau_{\text{F1,max}}$ denotes the
Messidor-2 decision threshold that maximises the F1 score, and the
corresponding precision, recall, and $F1_{\max}$ are reported at this operating point.
\textbf{Bold} values indicate the best result for each column.}
\label{tab:messidor2_results}

\begin{tabular}{ll|ccccc}
\toprule
\multicolumn{2}{c|}{Model}
  & AUC (\%)
  & Precision (\%)
  & Recall (\%)
  & $F1_{\max}$
  & Threshold $\tau_{\text{F1,max}}$\\
\cmidrule(lr){1-2} \cmidrule(lr){3-7}
Family & Architecture
  & 
  & 
  & 
  & 
  &  \\
\midrule
CNN    & ResNet50        & 66.8 & 42.2 & 51.2 & 0.47 & 0.092 \\
CNN    & DenseNet121     & 75.8 & 50.4 & 60.4 & 0.55 & 0.158 \\
CNN    & EfficientNetB3  & 79.7 & 59.8 & 61.5 & 0.61 & 0.286 \\
CNN    & InceptionV3     & 75.1 & 50.8 & 56.2 & 0.54 & 0.399 \\
\midrule
ViT    & CrossViTSmall   & 79.9 & 63.3 & 56.9 & 0.60 & 0.198 \\
ViT    & DeiTSmall       & 74.4 & 54.6 & 54.3 & 0.55 & 0.340 \\
ViT    & SwinTiny        & 79.7 & 71.4 & 54.7 & 0.62 & \textbf{0.508} \\
ViT    & ViTSmall        & 74.3 & 48.1 & 64.1 & 0.55 & 0.314 \\
\midrule
Hybrid & CoAtNet0        & 81.0 & 66.6 & 57.1 & 0.62 & 0.148 \\
Hybrid & MaxViTTiny      & \textbf{84.7} & \textbf{75.1} & 60.6 & 0.67 & 0.153 \\
\midrule
VLM    & CLIP ViT-B/16   & 81.0 & 69.3 & 56.2 & 0.62 & 0.072 \\
VLM    & SigLIP-Base384  & \textbf{84.7} & 69.7 & \textbf{67.0} & \textbf{0.69} & 0.250 \\
\bottomrule
\end{tabular}
\end{table*}

While RFMiD provides a comprehensive internal benchmark, a clinically useful screening system must also generalize to images acquired at different centers, using different devices and drawn from diverse patient populations. To assess out-of-domain robustness, we evaluate all RFMiD-trained models on the Messidor-2 dataset, a widely used benchmark for diabetic retinopathy (DR) screening. Messidor-2 contains macula-centered fundus photographs with adjudicated DR grades and gradability flags. In this study, we focus on the referable DR endpoint, defined as a Messidor-2 diagnosis $\geq 2$, which is clinically relevant for guiding referral decisions in screening settings.

For Messidor-2 evaluation, we apply each RFMiD-trained model checkpoint without any additional fine-tuning, keeping the full training pipeline unchanged. We use the diabetic retinopathy head from the multi-label classification model trained on RFMiD and compute threshold-free discrimination using area under the ROC curve (AUC). To summarize threshold-dependent performance, we select, for each model, the decision threshold that maximizes the F1 score ($\tau_{\text{F1,max}}$) and report the corresponding precision, recall, and $F1_{\max}$.

Table~\ref{tab:messidor2_results} summarizes referable DR performance on Messidor-2 for all architectures. Overall, AUC values range from 66.8\% (ResNet50) to 84.7\% (MaxViTTiny and SigLIP-Base384), indicating a clear performance drop compared with RFMiD, but preserved discriminatory ability across model families. Among CNN baselines, EfficientNetB3 achieves the best precision--recall trade-off at $\tau_{\text{F1,max}}$ (AUC 79.7\%, precision 59.8\%, recall 61.5\%), whereas ResNet50 shows the weakest generalization (AUC 66.8\%, $F1_{\max} \approx 0.50$). Vision transformer models achieve intermediate to strong performance: CrossViTSmall and SwinTiny reach AUCs of 79.9\% and 79.7\%, respectively, with $\tau_{\text{F1,max}}$ favoring higher precision for SwinTiny (71.4\%) and higher recall for ViTSmall (64.1\%). 

Hybrid backbones perform particularly well, with CoAtNet0 (AUC 81.0\%) and MaxViTTiny (AUC 84.7\%) both achieving $F1_{\max} \approx 0.60$--0.70. MaxViTTiny yields the highest precision (75.1\%) while maintaining a recall of 60.6\%. The vision–language models also show competitive external performance: CLIP ViT-B/16 attains an AUC of 81.0\% with balanced precision and recall (69.3\% and 56.2\%, respectively), and SigLIP-Base384 matches MaxViTTiny in AUC (84.7\%) while achieving the highest recall (67.0\%) and $F1_{\max} \approx 0.70$. Taken together, these results show that hybrid and transformer-based architectures retain a performance advantage over classical CNNs on Messidor-2, and that SigLIP-Base384 in particular combines strong discrimination with favorable precision--recall trade-offs for referable DR screening.

The observed performance degradation on Messidor-2 relative to RFMiD is likely due to both domain shift and label set shift. RFMiD is a multi-label, multi-disease dataset with 28 retinal pathology categories, whereas Messidor-2 provides a single DR grade per image and does not label other co-occurring conditions. Models trained to recognize a broad range of abnormalities on RFMiD may therefore learn decision boundaries that do not align perfectly with Messidor-2’s grading protocol, disease prevalence, or imaging characteristics. This label space mismatch, along with differences in cameras, image quality, and patient demographics, plausibly contributes to the reduced but still acceptable AUC and F1 values observed.

Interestingly, vision-language models-particularly SigLIP-Base384, appear relatively stronger on Messidor-2 than on RFMiD. One plausible explanation is that contrastive image–text pretraining provides robustness to changes in imaging conditions and population characteristics not explicitly represented during RFMiD fine-tuning. Another contributing factor may be that the referable DR endpoint represents a single, semantically coherent disease concept, well aligned with the global, concept-level representations learned by VLMs from natural language supervision (e.g., text prompts describing diabetic retinopathy). In contrast, RFMiD requires simultaneous classification of 28 diverse and sometimes rare conditions, a setting in which specialized vision transformers and hybrid architectures maintain a clearer advantage.

\subsection{Limitations and future work}

This study has some limitations. All experiments are conducted on a single primary training dataset (RFMiD) and one external benchmark (Messidor-2), which may not capture the full diversity of real-world acquisition conditions. We rely on standard fine-tuning strategies and a simple adaptation of vision–language models, without exploring more advanced prompting, multi-task learning, or explicit domain adaptation techniques. Finally, while RFMiD supports multi-label, multi-disease classification, our external validation focuses on a single disease endpoint (referable DR), which limits direct assessment of cross-domain generalization for other pathologies.

Future work could address these limitations by extending the evaluation to additional multicenter datasets, investigating more sophisticated adaptation strategies for vision–language and foundation models, and studying label harmonization across datasets with differing grading protocols.

\section{Conclusion}

In this work, we systematically compare twelve visual architectures for automated multi-disease retinal screening. Using standardized training configurations and a unified evaluation and calibration protocol, we assess performance on two clinically relevant tasks: (i) a binary screening task and (ii) a multi-label classification task. Across both settings, attention-based architectures, particularly SwinTiny and the hybrid CoAtNet0 and MaxViTTiny models, consistently outperform classical convolutional networks, while vision–language models deliver competitive but not superior performance on RFMiD. When evaluated on the external Messidor-2 dataset with a referable diabetic retinopathy endpoint, hybrid and transformer models again achieve the strongest overall results, and the SigLIP-Base384 vision–language model demonstrates relatively improved robustness under domain and label set shift.

This study makes three main contributions. First, we establish a unified and clinically motivated benchmarking framework for convolutional networks, vision transformers, hybrid CNN–transformer models, and vision–language models on RFMiD, using a consistent protocol for calibration, operating point selection, and metric computation. This enables a fair comparison across model families and provides practical guidance for model selection in retinal screening. Second, we provide a detailed analysis of multi-label classification performance, showing that transformer and hybrid backbones significantly improve macro and micro F1 scores compared with convolutional baselines, especially for rare retinal pathologies. Third, we extend the evaluation beyond RFMiD by conducting external validation on Messidor-2 for referable diabetic retinopathy, quantifying how different model families perform under domain and label set shift and highlighting the relatively strong external performance of SigLIP-Base384.

More broadly, our results suggest that attention-based vision architectures, particularly Swin-based and hybrid CNN–transformer designs, are strong candidates for real-world retinal screening systems when computational resources permit. Vision–language models offer a promising direction for robust, concept-level disease detection under distribution shift. For developers and decision-makers considering deployment of AI-enabled screening systems, the comparative evidence presented here can inform prioritization of architecture families and guide expectations regarding performance trade-offs between internal validation and external clinical settings.

\begin{acks}
This work was supported by Mitacs in partnership with Ebovir Biotechnologie Inc.
The authors gratefully acknowledge this financial and collaborative support.
\end{acks}


\bibliographystyle{ACM-Reference-Format}
\bibliography{bibfile}

\end{document}